\documentclass[%
 amsmath,amssymb,
]{article}

\usepackage{setspace}
\usepackage{changepage}

\usepackage[utf8]{inputenc} 
\usepackage[T1]{fontenc}    
\usepackage{hyperref}       
\usepackage{url}            
\usepackage{booktabs}       
\usepackage{amsfonts}       
\usepackage{nicefrac}       
\usepackage{graphicx}
\usepackage{caption}
\usepackage{subcaption}
\usepackage{multirow}
\usepackage{amsmath}
\usepackage{mathtools}
\usepackage{xcolor}

\title{In-field grape berries counting\\ for yield estimation using dilated CNNs}

\author{
L. Coviello $\quad$ M. Cristoforetti\thanks{Corresponding author} $\quad$ G. Jurman $\quad$ C. Furlanello\\ 
Fondazione Bruno Kessler\\
Trento, Italy\\
\texttt{ \{coviello,mcristofo,jurman,furlan\}@fbk.eu}
}

\begin{document}

\maketitle
\begin{abstract}
Digital technologies ignited a revolution in the agrifood domain known as precision agriculture: a main question for enabling precision agriculture at scale is if accurate product quality control can be made available at minimal cost, leveraging existing technologies and agronomists' skills. As a contribution along this direction we demonstrate a tool for accurate fruit yield estimation from smartphone cameras, by adapting Deep Learning algorithms originally developed for crowd counting.
\end{abstract}

\section{Introduction}
\label{sec:intro}
An interesting trend with high economic and societal impact that emerged in recent years is precision agriculture, namely the employment of digital technologies to better assess the conditions of agricultural fields and improve production processes. 
By adopting precision agriculture it is possible to increase productivity while reducing the amount of treatment on crops, eventually increasing availability of safer food at lower costs. 
This revolution is based on a systematic use of technology, including the widespread adoption of sensors, both in-field and in-lab for quality control processes.
In addition to the expensive and highly accurate instruments used in lab, sensors on portable devices are constantly being developed in precision agriculture to support quality control, to dramatically reduce costs and obtain results which are comparable to the ones obtained in labs with traditional technologies. 
One important and appealing opportunity for farmers is to employ the smartphone they already have and use in their daily activities, with the addition of ad hoc technologies that can help boost their productivity.

Here we focus on the application of Deep Learning to enable a precision agriculture approach using low cost cameras. In particular we demonstrate that using everyday technologies like smartphones, in combination with the adaptation of recent deep neural networks for crowd counting, will lead to a non-destructive yield estimation in the context of wine production, through an automatic estimate of the number of berries forming a grape bunch. This simplified approach can overcome the current procedure based on destructive sampling (cutting off and weighting a collection of grape bunches) to obtain a yield estimate.

Measuring grape weight is of high interest for wine producers also in view of quality control aspects, e.g. to decide whether to thin the cluster or defoliate the shoot. As the amount of nutrients present in the ground and transmitted to the grapes is substantially constant \cite{poni2006effects}, regulating the grape weight has a critical impact on wine quality. The standard procedure estimates yield as a function of the number of vines per surface unit $N_v$, the number of grape bunches per vine $N_b$ and the average weight of the bunch $P_b$, combined as follows to obtain the yield: 
\begin{equation}\label{eq:y1}
Y = N_v \cdot N_b \cdot P_b .	
\end{equation}

Clearly, the method has practical limitations in particular connected with the possibility of obtaining long term forecasts. In fact, the average weight of the clusters $P_b$ can be accurately determined only close to the harvest phase and estimation based only on historical data is difficult because the weight of the clusters can significantly change from year to year. For the varieties considered in this study the cluster's weights collected in the last five years by the CAVIT laboratory are presented in Tab.~\ref{tab:cluster_weight}. From there we see that there are cases where the variation through the years is close to $10\%$. Last but not least, this is basically a destructive sampling technique.  

\begin{table}[!h]
    \centering
	\begin{adjustwidth}{-.2in}{0in}

	\begin{tabular}{lrrrrrrr}
		\toprule
		{} &  2013 &  2014 &  2015 &  2016 &  2017 &  2018 &  \% mean dev. \\
		\midrule
		Chardonnay          &   170 &   184 &   176 &   172 &   172 &   208 &     0.06 \\
		Lagrein            &   280 &   279 &   325 &   265 &   259 &   264 &     0.06 \\
		Marzemino           &   308 &   311 &   336 &   326 &   350 &   318 &     0.04 \\
		Pinot Gris        &   164 &   177 &   181 &   141 &   167 &   205 &     0.09 \\
		Pinot Noir          &   149 &   174 &   159 &   155 &   158 &   175 &     0.05 \\
		Sauvignon Blanc     &   169 &   208 &   173 &   163 &   178 &   205 &     0.09 \\
		Traminer &   138 &   155 &   174 &   143 &   157 &   151 &     0.06 \\
		\bottomrule
	\end{tabular}
	\end{adjustwidth}
    \caption{Average  cluster weight for different grape varieties in Trentino (Italy) for the five years between 2013 and 2018.}
    \label{tab:cluster_weight}	
\end{table}

In Tab.~\ref{tab:berry_weight} the average weight of single berries is reported: comparing Tab.~\ref{tab:cluster_weight} with Tab.~\ref{tab:berry_weight} we can see that in most of the cases, the average berry's weight is more stable through the years than cluster's weight. This suggests that combining the historical series of berry's weight with accurate berry counting we can deliver better results than using clusters weight. Moreover the use of the historical data opens the possibility to have a yield estimate immediately after the fruit sets. 

Following this approach the Eq.~\ref{eq:y1} becomes:
\begin{equation}\label{eq:y2}
Y = N_v \cdot N_b \cdot N_a \cdot P_a,
\end{equation}
with $N_a$ the average number of berries per bunch and $P_a$ the average berry's weight.

\begin{table}[!h]
    \centering
	\begin{tabular}{lrrrr}
		\toprule
		{} &  2016 &  2017 &  2018 &  \% mean dev. \\
		\midrule
		Chardonnay          &   1.6 &   1.6 &   1.7 &     0.03 \\
		Lagrein             &   1.9 &   2.2 &   2.0 &     0.06 \\
		Marzemino          &   2.1 &   2.3 &   - &     0.05 \\
		Pinot Gris        &   1.4 &   1.6 &   1.6 &     0.06 \\
		Pinot Noir          &   1.5 &   1.6 &   1.6 &     0.03 \\
		Sauvignon Blanc     &   - &   1.8 &   1.6 &     0.06 \\
		Traminer  &   1.4 &   1.7 &   1.7 &     0.08 \\
		\bottomrule
	\end{tabular}
    \caption{Average single berry weight for different grape varieties in Trentino (Italy) for the years 2016, 2017, 2018.}
    \label{tab:berry_weight}
	
\end{table}

In this work we look for non-destructive approaches for grape yield estimation, applicable immediately after the fruit set. The solution proposed is based on the use of images with smartphones and application of deep learning algorithms to count the number of berries in the images. 

Counting is the core step for yield estimation for fruit; for grapes, automatic image analysis used 3D bunch reconstruction or artificial illumination at night \cite{Liu2015, Font2015}, while other Android based solutions used a capturing box as a synthetic background \cite{Aquino2018}. 

The peculiarity of the method presented in this work is that in our framework there is no need of particular preparation for image acquisition, enabling an easier and faster AI-based yield estimation system. This opens the possibility of testing two different strategies for the yield estimation: the first is based on the evaluation of the average number of grape for bunch to evaluate Eq.~\ref{eq:y2}. The second is having a picture of the whole grape field (for example as a panoramic view), estimate the total number of berries and then simply multiply this for the average berry's weight.
The input of the networks for the two methods are images with slightly different characteristics. In Sec.~\ref{sec:results} we show the results obtained on datasets optimised for the two different approaches.

\section{From crowd to berries counting}

\begin{figure}[!t]
    \centering
    \begin{subfigure}{0.32\textwidth}
        \includegraphics[width=.95\linewidth]{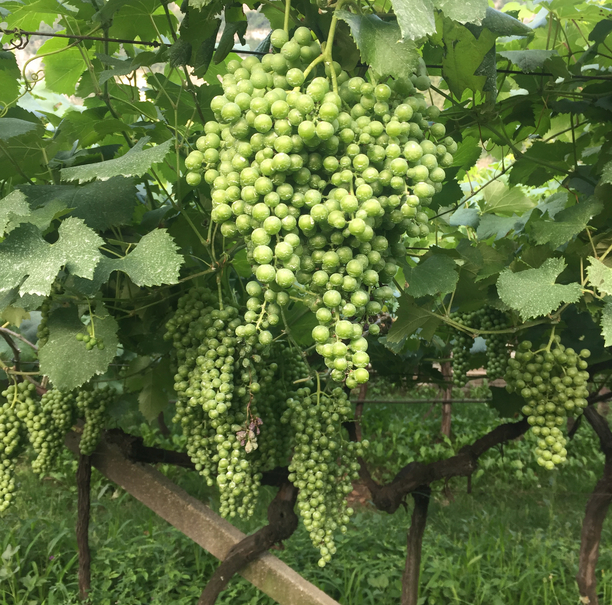}
        \caption{Input sample}
        \label{fig:grape_sample_best_output_1}
    \end{subfigure}
    \begin{subfigure}{.32\textwidth}
        \includegraphics[width=.95\linewidth]{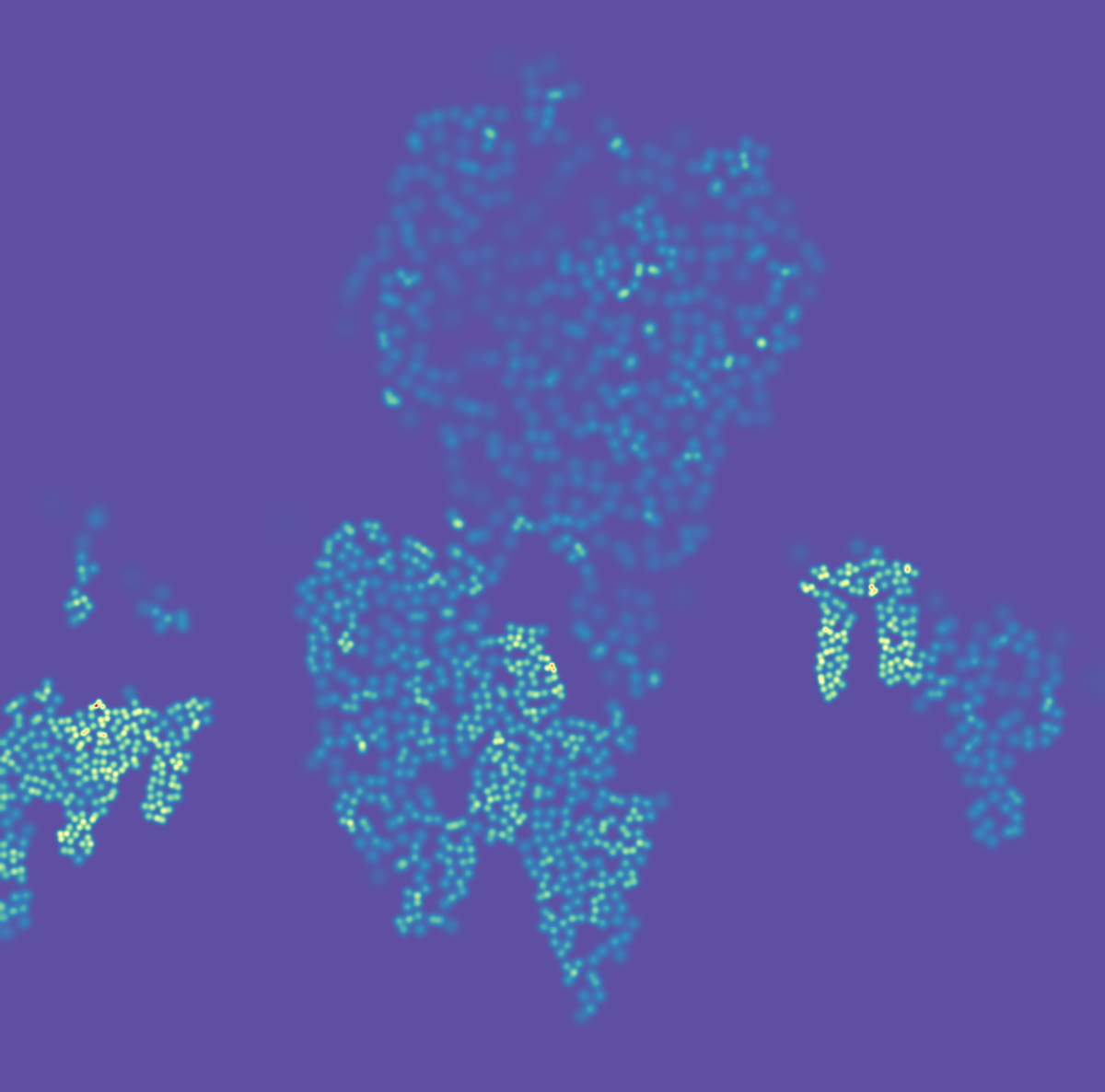}
        \caption{Ground truth}
        \label{fig:grape_sample_best_output_2}
    \end{subfigure}
    \begin{subfigure}{.32\textwidth}
        \includegraphics[width=.95\linewidth]{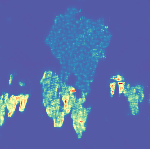}
        \caption{Original prediction}
        \label{fig:grape_sample_best_output_3}
    \end{subfigure}
    \caption{Example of application of CSRNet on a CR2 image (\subref{fig:grape_sample_best_output_1}), its associated ground truth (\subref{fig:grape_sample_best_output_2}) and model output (\subref{fig:grape_sample_best_output_3}).}
    \label{fig:grape_sample_best_output}
\end{figure}

At the core of the proposed method for yield estimation is the ability of accurate automatic counting of berries from pictures taken in the grape fields. We will show that for these task, techniques developed in the context of automatic crowd counting can successfully be adapted: in the case of congested scene recognition presented in \cite{li2018csrnet} the input picture is processed by a Deep Neural Network  (CSRNet) that  returns a density map, the integral of which is the estimated amount of subjects to count, in our case the number of berries in the image (Fig.~\ref{fig:grape_sample_best_output}).

The CSRNet architecture employs the first ten convolutional layers of VGG16 \cite{simonyan2014very} pretrained on ImageNet \cite{deng2009imagenet} as feature extractor and a dilated CNN for density map generation. Training from scratch the full network requires an enormous amount of annotated data, and annotation is an expensive operation in particular with grape images where labelling is required at the level of single berry. To reduce the number of annotations required for the training we adopt a transfer learning approach where a pre-trained VGG16 model is used as a generalized feature extractor for the training of the the last part of the network. The use of dilated convolutions, i.e. convolutions with non contiguous kernels with a larger receptive field, aggregates multi-scale contextual information while maintaining the same spatial resolution.

The training phase is based on the generation of density maps as ground truth. This requires the annotation of the images at single berry level: given an input image, a berry at the position $x_i$ is represented as a Dirac delta function $\delta(x - x_i)$ which represents a binary mask with only the point $x_i$ set to 1.
After the annotation the image is represented as:

\begin{equation*}
 H(x) = \sum_{i=1}^{N}\delta(x - x_i),
\end{equation*}
where $N$ is the number of labeled points. 
To obtain a continuous density function $F(x)$ from the discrete representation $H(x)$ \cite{zhang2016single} we use a convolution with a Gaussian kernel $G_\sigma$ using $F(x) = H(x) * G_\sigma(x)$ as introduced in \cite{lempitsky2010learning}, where the $\sigma$ fixes the level of smoothing in the mask.
Additionally, to tackle the presence of dense scenes in the images we use geometry-adaptive kernels \cite{zhang2016single} which consider how the neighbours of a labeled point are distributed. Geometry-adaptive kernels are defined as:

\begin{equation*}
 F(x) = \sum_{i=1}^N\delta(x - x_i) * G_{\sigma_i}(x), \textrm{ with } \sigma_i = \beta \overline{d_i}, 
\end{equation*}
where $\overline{d_i}$ is the average distance of the $k$ nearest neighbours of $x_i$ and $\beta$ is a regularization parameter. In all the experiments we use the same configuration of \cite{zhang2016single}, setting $k = 3$ and $\beta = 0.3$.

\section{In-field images}
\label{sect:dataset}

\begin{figure}[!t]
    \centering
    \begin{subfigure}{0.63\textwidth}
        \includegraphics[width=.95\linewidth]{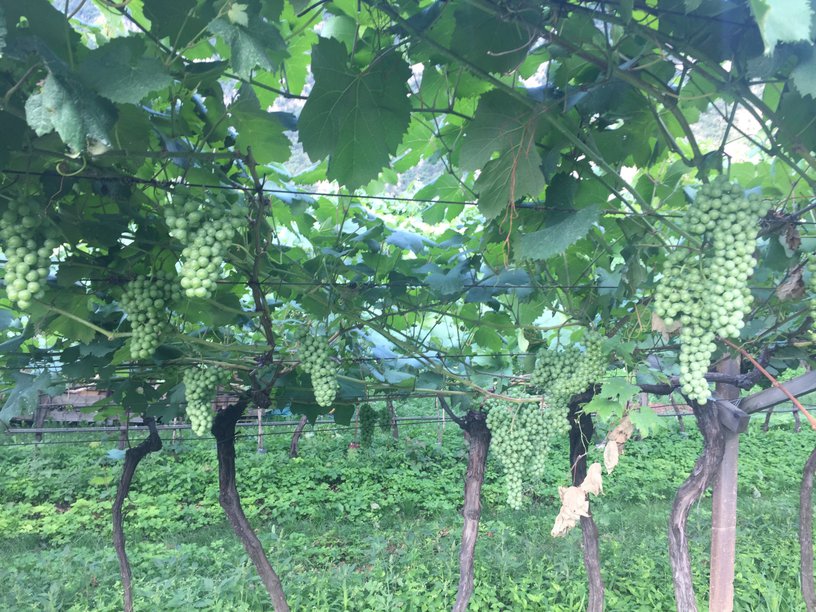}
        \caption{CR1 example}
        \label{fig:grape_ds_1}
    \end{subfigure}
    \begin{subfigure}{.35\textwidth}
        \includegraphics[width=.95\linewidth]{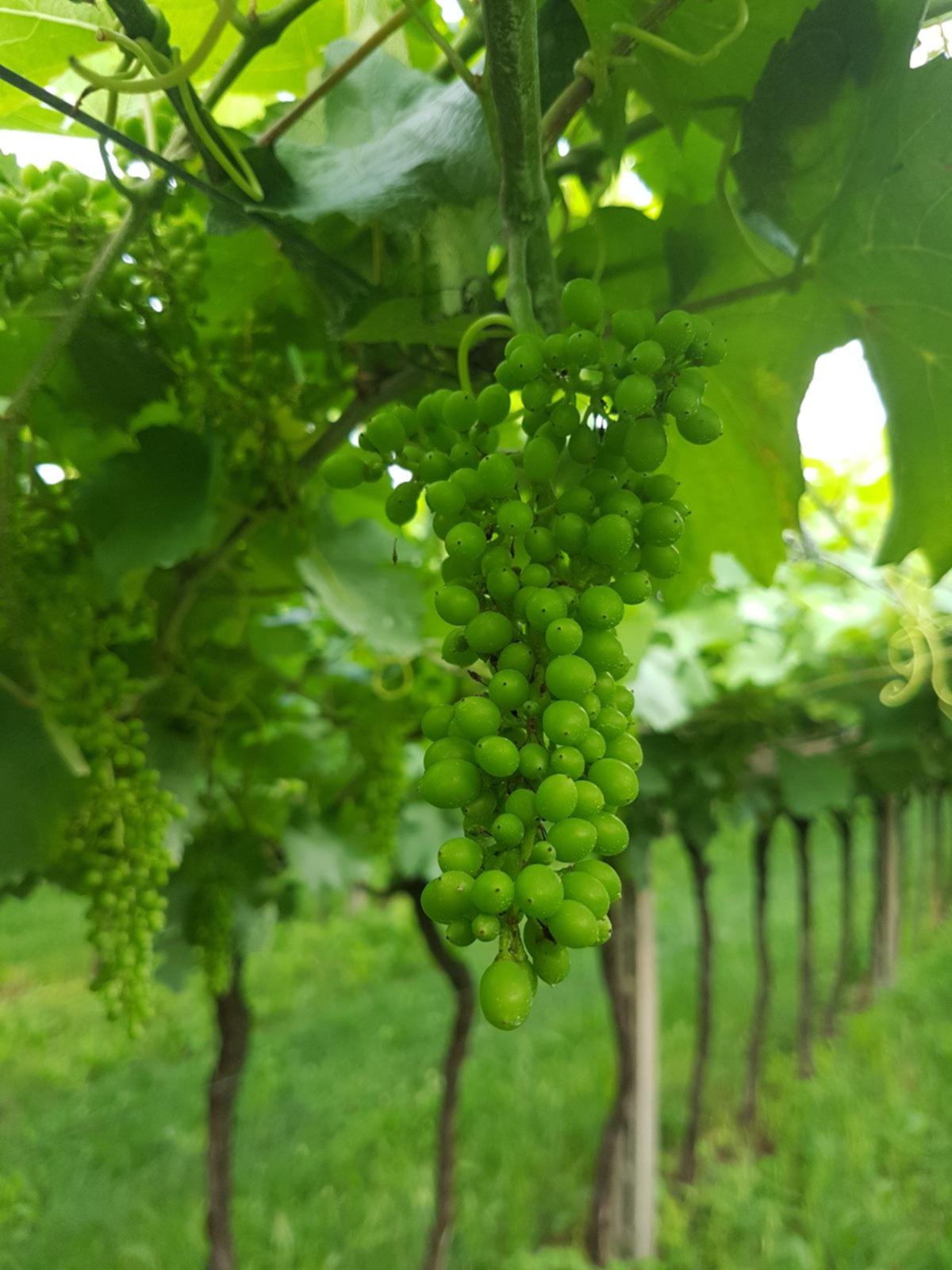}
        \caption{CR2 example}
        \label{fig:grape_ds_2}
    \end{subfigure}
    \caption{Example of medium distance and close-up images present in CR1 (\subref{fig:grape_ds_1}) and CR2 (\subref{fig:grape_ds_2}) datasets respectively.}
    \label{fig:grape_ds}
\end{figure}

The performances of the proposed methodology are tested on two datasets (Tab.~\ref{tab:datastats}). The first one is composed of 128 close-up and manually labeled images belonging to 7 different varieties (Tab.~\ref{tab:cr1stats}), taken with  8Mpx and 2Mpx smartphone cameras from which we extracted 17,006 single berry annotations (CR1 dataset). The second one is composed of 18,865 manually labeled single berry annotations (CR2 dataset), derived from 17 images of the Teroldego variety taken with a smartphone camera at 8Mpx resolution (2448$\times$3268 pixels) from a medium distance (1-1.5~m). Examples for the images in the two datasets are presented in Fig~\ref{fig:grape_ds}.

\begin{table}[!ht]
    \centering
    \begin{tabular}{lccccc}
        \toprule
        \textbf{Dataset}  & \textbf{Images} & \textbf{Max} & \textbf{Min} & \textbf{Avg} & \textbf{Total} \\ 
        \midrule
        CR1      & 128 & 322 & 42  & 132.9  & 17006 \\        
        CR2      & 17 & 1789 & 543 & 1109.7 & 18865 \\
        \bottomrule
    \end{tabular}
    \caption{Berries and bunches annotations in the CR1 and CR2 datasets.}
    \label{tab:datastats}
\end{table}

\begin{table}[!ht]
    \centering
    \begin{tabular}{lrrrrr}
        \toprule
        Variety &  Images &  Min &  Max &        Mean &  Total \\
        \midrule
        Chardonnay   &       7 &  172 &   51 &  104.71 &    733 \\
        Lagrein      &       9 &  211 &  117 &  163.22 &   1469 \\
        Marzemino    &      16 &  244 &   53 &  114.81 &   1837 \\
        Pinot Gris   &      34 &  322 &   86 &  150.91 &   5131 \\
        Pinot Noir   &      21 &  269 &   93 &  142.00 &   2982 \\
        Sauvignon    &      21 &  167 &   42 &  110.38 &   2318 \\
        Traminer     &      20 &  207 &   61 &  126.80 &   2536 \\
        \bottomrule
    \end{tabular}
    \caption{CR1 dataset description by variety.}
    \label{tab:cr1stats}
\end{table}

The CR1 images were taken in a stage where berries are still small and well separated, therefore clusters are characterized by a low degree of occlusion. In addition, the dataset was collected trying to include only one bunch in every picture. For the evaluation of the CSRNet performance the dataset was randomly split in 102 images for train and 26 images for test, corresponding to 13,353 berries in training and 3,653 berries in test. 5-fold cross validation was applied on the training dataset.

In the CR2 dataset each image contains more than one cluster, with different sizes both in the foreground and in the background. The images are randomly split in 11 images for train and 6 images for validation, corresponding to an average of 12,577 berries in training and 6,288 in validation, respectively. In this case 3-fold cross validation was applied.

For both datasets, given that the environment where the pictures are taken is not controlled, there is a large variance between images under different aspects. First, the clusters are visually very different in brightness and saturation while there is little difference with the colors between grapes and the surrounding leaves. 
This represents a challenge given that intra class variance (e.g. colors between bunches) is higher than extra class variance (e.g. bunches versus leaves). 

For CR2 an additional challenge is given by the clusters main dimension, that may range from 1000px (around 40\% of the total height with the landscape orientation) of the image to 70px (0.03\%). Finally, the CR2 dataset is characterized by images of grapes before veraison, in a stage where berries are almost of the final size, presenting a high degree of occlusion between berries, increasing the difficulty of the task.

The preprocessing techniques adopted for all the experiments are channel normalization as per VGG-16 and the resizing of the images at 800px height. For the training phase of the CSRNet model we crop from the input image a patch at a random location every time with 1/4 of its original size.
On top, we use standard data augmentation techniques in order to extend the dataset for the training phase of the model.

\section{Yield estimation}
\label{sec:results}

As explained in Sec.~\ref{sec:intro} we explore two different strategies for yield estimation using deep learning. The first one based on Eq.~\ref{eq:y2} has images taken at small distances with only one grape bunch on focus, the second considers panoramic images collected from a distance of 1-2m that potentially can capture a wide portion of the field (in the order of thousands of berries).

We present here berries counting performances of CSRNet on the two datasets CR1 and CR2 as test of the feasibility of the two approaches. In all the experiments we employed the Adam optimizer, setting the initial learning rate as $10^{-5}$ and $10^{-4}$ for CR1 and CR2 respectively, dropping the learning rate by an order of magnitude every 50 epochs. Considering the small amount of images of the two datasets, we froze the feature extraction layers (i.e. first ten VGG-16 layers) and updated only the dilated CNN layer weights for density map generation. With this approach, all the training processes converged in less than 200 epochs, and we evaluated the performances of CSRNet using the weights of the last training epoch. Finally the number of patches for each iteration (i.e. batch size) was set to 20 for CR1 and 4 for CR2, given the memory restrictions on the machine used for training and the larger size of CR2 images. For each patch there are an average of 71 berries for CR1 and 427 berries for CR2.

To evaluate the CSRNet model performance we use Mean Absolute Error (MAE) and Mean Squared Error (MSE) as in the crowd counting domain, which are defined as:

\begin{eqnarray*}
    &\textit{MAE} = \frac{1}{N} \sum_{i=1}^N \lvert C_i - C_i^{GT} \rvert, \,\,\,
    \textit{MSE} = \sqrt{\frac{1}{N} \sum_{i=1}^N ( C_i - C_i^{GT} )^2}&
\end{eqnarray*}

where $C_i$ is the estimated count and $C_i^{GT}$ is the ground truth count associated to image $i$. The estimated count is equal to the integral of the output density map.
These two metrics represent a measure of accuracy (MAE) and robustness (MSE) of the model.

To estimate crop yield it is important to consider the performances obtained when considering the cumulative sum of the outputs and ground truths as well. To this end, we also employ Overall MAE, defined as

$$ \textit{Overall MAE} = \lvert \sum_{i=1}^N C_i - \sum_{i=1}^N C_i^{GT} \rvert $$

and that gives information on the performances that can be obtained in practical applications of the system.

\begin{table}[!ht]
\centering
\begin{adjustwidth}{-.25in}{0in}

    \begin{tabular}{ll cccc}
        \toprule
                 &     & n &    MAE & MAE (\%) &    MSE \\
        \midrule
        \multirow{2}{*}{5-fold CV} & Per Image &    20.4 &   13.66 $\pm$ 4.70 &  11.16\% $\pm$ 2.70\% &  18.33 $\pm$ 6.33 \\
                                   & Overall   &  2670.6 &  56.48 $\pm$ 60.08 &   2.13\% $\pm$ 1.97\% &      \\
        \midrule
        \multirow{2}{*}{Test} & Per Image &    26 &  13.25 &  10.32\% &  16.07 \\
                              & Overall   &  3653 &  10.65 &   0.29\% &        \\
        \bottomrule
    \end{tabular}
\end{adjustwidth}    
    \caption{Application of CSRNet on CR1 5-fold CV and test sets. The N column refer to the average number of images and berries per fold and in the test set.}
    \label{tab:experiments_cr1}
\end{table}

In Tab. \ref{tab:experiments_cr1} results for CSRNet on CR1 are presented both for 5-fold cross validation and test. We report both the error per image and the overall error. The last is important in the assumption of having a unique grape bunch in the picture and being interested in the average number of berries per bunch: considering the full dataset helps averaging the over/under-estimation of the network on the single images. It is quite impressive to observe the drop in the percentage error when considering the whole dataset from $10\%$ to less than $1\%$ in test, showing the importance on averaging on many pictures.

\begin{figure}[!t]
    \centering
    \begin{subfigure}{0.32\textwidth}
        \includegraphics[width=.95\linewidth]{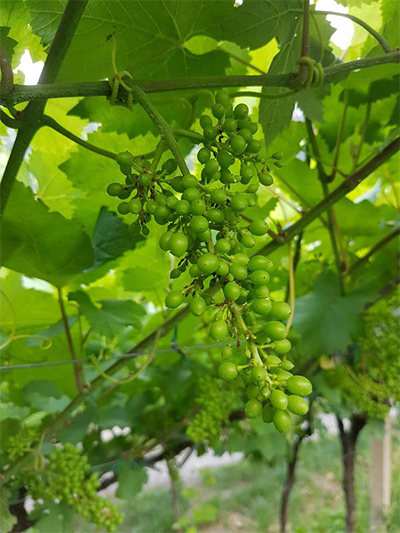}
        \caption{Input sample}
        \label{fig:grape_sample_focus_effect_1}
    \end{subfigure}
    \begin{subfigure}{.32\textwidth}
        \includegraphics[width=.95\linewidth]{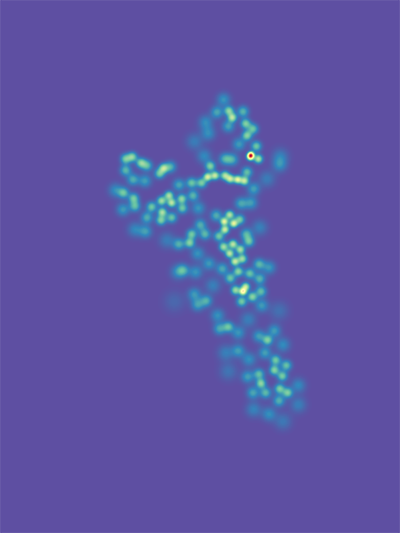}
        \caption{Ground truth}
        \label{fig:grape_sample_focus_effect_2}
    \end{subfigure}
    \begin{subfigure}{.32\textwidth}
        \includegraphics[width=.95\linewidth]{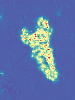}
        \caption{Original prediction}
        \label{fig:grape_sample_focus_effect_3}
    \end{subfigure}
    \caption{Example of application of CSRNet on a CR1 image (\subref{fig:grape_sample_best_output_1}), its associated ground truth (\subref{fig:grape_sample_best_output_2}) and model output (\subref{fig:grape_sample_best_output_3}).}
    \label{fig:grape_sample_focus_effect}
\end{figure}

An interesting aspect of the network behaviour is related to having a single cluster on focus in the CR1 dataset. In fact even though the images present a high level of blur in the background due to the closeness of the camera to the main clusters, smaller bunches are still present in the background. However given that only the foremost berries were labeled in CR1 images, the network learns to ignore the background clusters and focuses only to the one present in the foreground using as discriminant features sharpness and sizes of berries edges (Fig.~ \ref{fig:grape_sample_focus_effect}).

\begin{figure}[!t]
    \centering
    \begin{subfigure}{0.32\textwidth}
        \includegraphics[width=.95\linewidth]{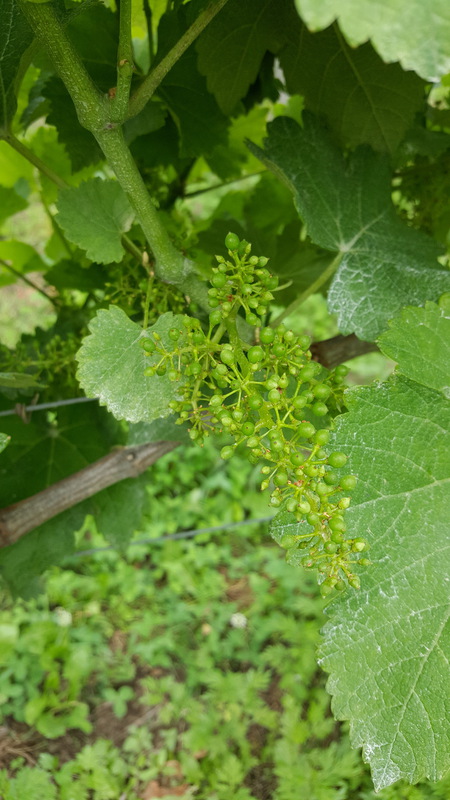}
        \caption{Sauvignon}
        \label{fig:pheno_diff_1}
    \end{subfigure}
    \begin{subfigure}{.32\textwidth}
        \includegraphics[width=.95\linewidth]{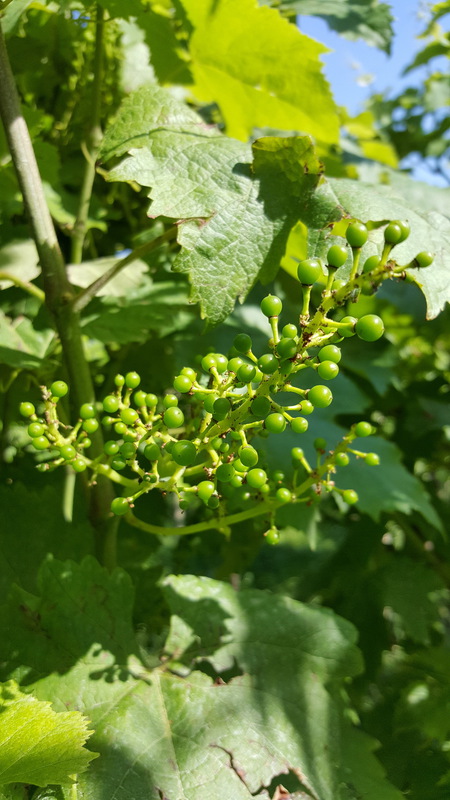}
        \caption{Marzemino}
        \label{fig:pheno_diff_2}
    \end{subfigure}
    \begin{subfigure}{.32\textwidth}
        \includegraphics[width=.95\linewidth]{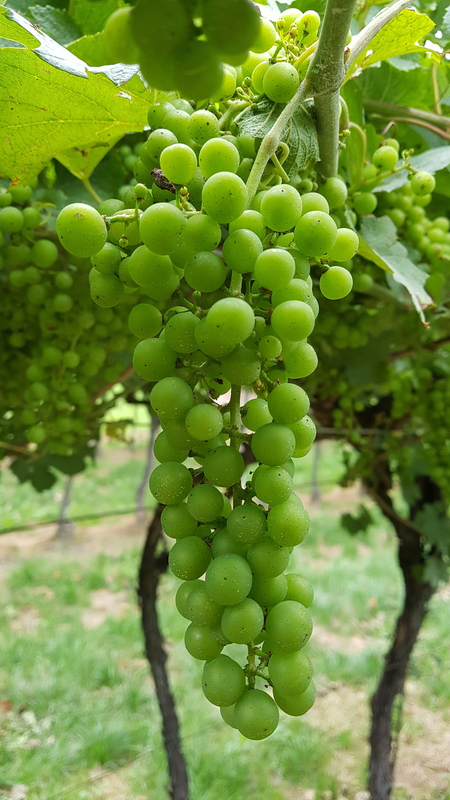}
        \caption{Pinot Gris}
        \label{fig:pheno_diff_3}
    \end{subfigure}
    \caption{CR1 images are collected during the same time period but with different at phenological stages depending on the variety. Pictures are sorted by development stage (from less to most developed from left to right).}
    \label{fig:pheno_diff}
\end{figure}

The CR1 dataset collects pictures of seven different varieties, the performances of the network for each variety is reported in Tab.~\ref{tab:experiments_cr1_var_val} and Tab.~\ref{tab:experiments_cr1_var_test}. The difference in performance reflects that having collected the pictures in the same days for all the varieties implies a non uniform phenological state, that implies highly different visual features (Fig.~ \ref{fig:pheno_diff}). Although this difference among varieties impacts the performances of CSRNet on single image prediction,  the model is capable to obtain a low MAE when aggregating the output predictions for almost all the varieties.

\begin{table}[!ht]
\begin{adjustwidth}{-1.25in}{0in}

\centering
    \begin{tabular}{l|cccc|ccc}
        \toprule
            {} & \multicolumn{4}{c|}{Per Image} & \multicolumn{3}{c}{Overall} \\
            {} & n                 &        MAE & MAE (\%) &    MSE &  N & MAE & MAE (\%) \\
        \midrule
            Chardonnay &  1.0 &    4.69 $\pm$ 3.53 &    4.23\% $\pm$ 2.87\% &    4.69 $\pm$ 3.53 &  112.8 &    4.69 $\pm$ 3.53 &   4.23\% $\pm$ 2.87\% \\
            Lagrein    &  1.4 &    5.41 $\pm$ 3.23 &    3.36\% $\pm$ 1.79\% &    5.61 $\pm$ 3.51 &  228.2 &    4.63 $\pm$ 3.22 &   2.29\% $\pm$ 1.68\% \\
            Marzemino  &  2.6 &  18.48 $\pm$ 17.43 &  16.29\% $\pm$ 11.48\% &  21.29 $\pm$ 18.71 &  307.2 &  19.20 $\pm$ 16.00 &  8.78\% $\pm$ 10.02\% \\
            Pinot Gris &  5.4 &    9.57 $\pm$ 3.75 &    6.84\% $\pm$ 3.10\% &   11.59 $\pm$ 4.39 &  766.6 &  36.60 $\pm$ 27.58 &   4.60\% $\pm$ 3.50\% \\
            Pinot Noir &  3.4 &   14.30 $\pm$ 8.48 &   10.88\% $\pm$ 5.87\% &   16.08 $\pm$ 9.33 &  480.0 &  16.37 $\pm$ 13.82 &   3.68\% $\pm$ 3.64\% \\
            Sauvignon  &  3.4 &   21.35 $\pm$ 7.03 &   19.33\% $\pm$ 6.22\% &   25.08 $\pm$ 8.51 &  367.0 &  50.55 $\pm$ 28.15 &  13.88\% $\pm$ 6.77\% \\
            Traminer   &  3.2 &  14.02 $\pm$ 11.28 &   11.52\% $\pm$ 9.18\% &  15.88 $\pm$ 12.72 &  408.8 &  24.02 $\pm$ 32.51 &   4.95\% $\pm$ 5.60\% \\
        \bottomrule
    \end{tabular}
\end{adjustwidth}
    \caption{Application of CSRNet on CR1 dataset with 5-fold CV. Results are reported for all the varieties used in this work. The N columns refer to the average number of images and berries per fold respectively.}
    \label{tab:experiments_cr1_var_val}
\end{table}

\begin{table}[!ht]
\centering
    \begin{tabular}{l|cccc|ccc|}
        \toprule
            {} & \multicolumn{4}{c}{Per Image} & \multicolumn{3}{c}{Overall} \\
            {}         &  n &       MAE & MAE (\%) &    MSE &     N &     MAE & MAE (\%) \\
        \midrule
            Chardonnay &  2 &      7.74 &   8.79\% &   8.38 &   169 &    6.38 &   3.77\% \\
            Lagrein    &  2 &     11.03 &   6.94\% &  11.88 &   328 &   22.05 &   6.72\% \\
            Marzemino  &  3 &     13.77 &  14.32\% &  16.99 &   301 &   35.31 &  11.73\% \\
            Pinot Gris &  7 &     19.86 &  13.00\% &  22.98 &  1298 &   11.08 &   0.85\% \\
            Pinot Noir &  4 &     10.36 &   7.80\% &  10.91 &   582 &   11.62 &   2.00\% \\
            Sauvignon  &  4 &     10.35 &   8.62\% &  12.79 &   483 &   12.54 &   2.60\% \\
            Traminer   &  4 &     10.95 &   9.31\% &  12.22 &   492 &    5.52 &   1.12\% \\
        \bottomrule
    \end{tabular}
    \caption{Application of CSRNet on CR1 test dataset. Results are reported for each variety. The N columns refer to the number of images and berries in the test set.}
    \label{tab:experiments_cr1_var_test}
\end{table}

\begin{table}[!t]
\centering
    \begin{tabular}{lcccc}
        \toprule
        {} &        n   &       MAE &       MAE (\%) &             MSE \\
        \midrule
        Per Image &     5.7 &   117.36 $\pm$ 14.07 &  10.74 $\pm$ 1.15 &  137.81 $\pm$ 18.19 \\
        Overall   &  6288.3 &  466.53 $\pm$ 182.99 &   7.24 $\pm$ 1.53 &                  \\
        \bottomrule
    \end{tabular}
    \caption{Application of CSRNet on CR2 dataset with 3-fold CV. The N columns refer to the average number of images and berries per fold respectively.}
    \label{tab:experiments_cr2}
\end{table}

Tab. \ref{tab:experiments_cr2} collects results for CSRNet tested on CR2 dataset with 3-fold cross validation. Considering single images predictions with an average of 1113.9 berries per picture, the model reaches a MAE of 117.36 berries for each validation fold (10.74\%). The overall MAE obtained comparing the cumulative sum of predictions and ground truth (6288.3 berries in average for each fold) results in lower value, i.e. 466.53 (7.24\%), benefiting from the balancing effect of over- and underestimation when aggregating predictions.

This result is fully comparable with what obtained with the alternative methods available in the literature, but where the images are taken in a controlled environment or employing a capturing box to limit background interference.

\section{Conclusions and future work}
We demonstrate that crop yield estimation for grape berries can be obtained from medium smartphone quality by an adaptation of algorithms for crowd counting. In particular we found promising results in the challenging case of data collected directly in field and without special cautions or the employment of additional tools. 

Further research will investigate the problem of estimating weight clusters from in field images, which depends on the real amount of berries present on a cluster and not only on those that can be visually counted from a picture.


\subsubsection*{Acknowledgments}
Authors thank CAVIT s.c. for providing data and scientific support throughout all phases of the research.
LC is supported by Consortium GARR (https://www.garr.it) under the grant "Orio Carlini" scholarship for young graduates.

\bibliographystyle{unsrt}
\small
\bibliography{gr_count}
\end{document}